%% file: iclr2027_conference.tex
\documentclass{article} 
\usepackage{iclr2027_conference,times}

\input{math_commands.tex}

\usepackage{hyperref}
\usepackage{url}
\usepackage{booktabs}
\usepackage{graphicx}
\usepackage{multirow}
\usepackage{fancyhdr}
\usepackage{tikz}

\renewcommand{\headrulewidth}{0pt}

\AddToHook{shipout/foreground}{%
  \begin{tikzpicture}[remember picture, overlay]
    \node[anchor=south, font=\small, text=red]
      at ([yshift=15mm]current page.south)
      {This manuscript is under review. Please contact yshi457@gatech.edu for up-to-date information};
  \end{tikzpicture}%
}

\newcommand{\projectname}[1]{\textsc{Rondo}}

\title{\projectname{}: Unsupervised Discovery of Recurring Temporal Structure}

\author{
Yingtian Shi\thanks{Both authors contributed equally to this research.}\\
School of Interactive Computing\\
Georgia Institute of Technology\\
\texttt{yshi457@gatech.edu} \\ 
\And
Ankith Chandra\footnotemark[1] \\
School of Interactive Computing\\
Georgia Institute of Technology\\
\texttt{ankithnchandra@gatech.edu} \\
\And
Thomas Plötz \\
School of Interactive Computing\\
Georgia Institute of Technology\\
\texttt{thomas.ploetz@gatech.edu}
}

\iclrfinalcopy 
\usepackage{float}
\begin{document}

\maketitle

\begin{abstract}
Many real-world time series data exhibit structural properties at multiple scales, from short, recurring units to complex sequences composed of these units. Unsupervised discovery of both these components and structure enables the design of intelligent systems that help interpreting temporal data thereby limiting the amount of costly human annotations required. 
Existing modeling approaches typically overlook the hierarchical structure inherent to many time series, treating recurring patterns at different temporal scales as independent structures. Moreover, most assume access to the complete data sequence and treat discovery as a static process, limiting their ability to evolve as the stream grows. 

We introduce \projectname{}, an unsupervised approach for modeling recurring hierarchical structure in continuous temporal streams. By explicitly constructing vocabularies of reusable units and their recurring compositions, \projectname{} captures structure shared across complex temporal patterns while refining and expanding its discoveries as the stream evolves. 
Evaluations on temporal sequences spanning diverse domains and data modalities show that \projectname{} outperforms existing unsupervised recurrence-discovery baselines, with particularly pronounced advantages in limited-data and continual-stream settings.
These capabilities support recurring-pattern discovery, scalable behavior understanding, and adaptive intelligent systems operating on long, unlabeled temporal streams.
\end{abstract}

\input{sections/intro}
\input{sections/relatedwork}

\input{sections/method}
\input{sections/experiment}

\section{Conclusion}
We introduced \projectname{}, an unsupervised approach

for discovering multiscale recurring structure in continuous temporal streams. By explicitly modeling reusable local units and their higher-level compositions, \projectname{} captures recurring structure beyond independent states or subsequences and continually updates its vocabularies as the stream evolves. Experiments across diverse datasets demonstrate improved discovery and temporal segmentation over existing unsupervised discovery methods. These results highlight the benefits of modeling temporal recurrence as compositional and evolving structure, and motivate extending it to richer temporal hierarchies in future work.

\subsection*{AI use statement}

In this work, we used generative AI tools to assist with dataset preprocessing and reformatting, and with implementing and debugging code for baseline experiments. This included adapting released baseline implementations to our data format, writing evaluation and scoring scripts, and developing scripts for running and aggregating experiments. We also used generative AI tools to improve the clarity and readability of portions of the manuscript.

We did not use generative AI tools to design the proposed method, formulate research hypotheses, design the experimental methodology, interpret experimental results, or generate research ideas. 

We reviewed and verified all AI-assisted work before inclusion in the paper. In particular, reported results were cross-checked against saved configurations and per-recording outputs, baseline settings were checked against the corresponding released implementations, and modifications to released baseline implementations are documented in the appendix. All AI-assisted text was reviewed and edited by the authors. 

\subsection*{Ethics statement}
This work uses publicly available datasets and does not involve new data collection or human-subject studies. We follow the usage terms of the respective datasets and use the data solely for research purposes. Our method performs unsupervised temporal structure discovery and does not infer sensitive personal attributes. We are not aware of additional ethical concerns specific to this work beyond those associated with the original datasets and their collection protocols.

\subsection*{Reproducibility statement}
We provide detailed descriptions of the proposed method, preprocessing pipeline, evaluation protocol, and experimental settings in the main paper and appendix. The appendix includes dataset-specific preprocessing and windowing settings, method hyperparameters, metric definitions, implementation details, and all modifications made to released baseline implementations. We also report results across multiple random seeds where applicable. Code and scripts for reproducing the experiments will be released following the review process.

\bibliography{iclr2027_conference}
\bibliographystyle{iclr2027_conference}

\appendix
\section*{Appendix}
\input{sections/app_qual}

\input{sections/app_metrics}

\input{sections/app_dataset}

\input{sections/app_method}
\input{sections/app_baselines}

\input{sections/app_results}

\end{document}

%% file: math_commands.tex
\usepackage{amsmath,amsfonts,bm}

\def\eqref#1{equation~\ref{#1}}

\def\1{\bm{1}}

\DeclareMathAlphabet{\mathsfit}{\encodingdefault}{\sfdefault}{m}{sl}
\SetMathAlphabet{\mathsfit}{bold}{\encodingdefault}{\sfdefault}{bx}{n}



%% file: sections/intro.tex
\section{Introduction}
\label{sec:intro}

\begin{figure}[htbp]
    \centering
    \includegraphics[width=\linewidth]{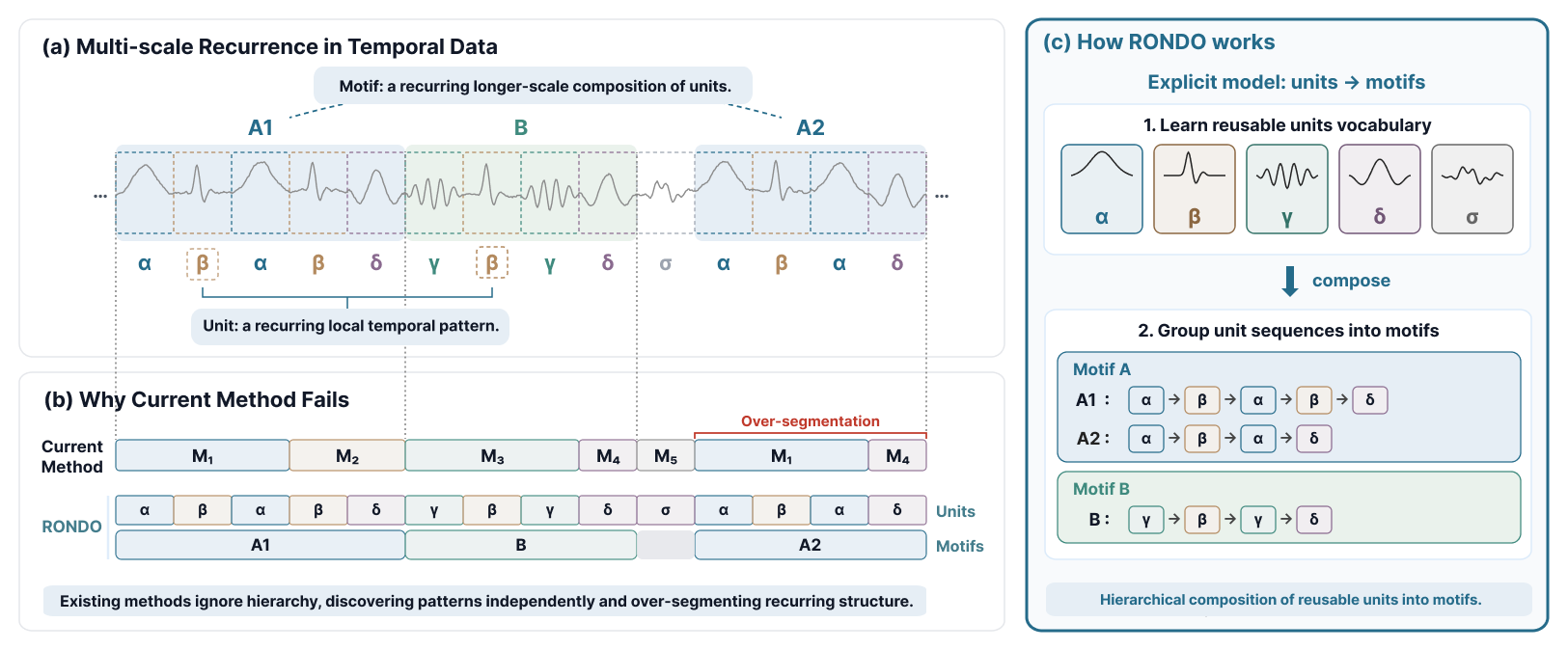}
    \caption{
{RONDO discovers recurring temporal structure hierarchically.}
(a) Temporal streams exhibit recurring local units that compose into longer-scale motifs.
(b) Current methods can fragment such structure and cause over-segmentation, while \projectname{} preserves the unit-motif hierarchy.
(c) \projectname{} learns reusable units and composes recurring unit sequences into motifs.
}
    \label{fig:teaser}
\end{figure}

Continuous temporal data often exhibit recurring structure across multiple temporal scales. 
At a local scale, subsequences that differ in their signals, duration, or surrounding context may still represent the same underlying temporal unit~\citep{zhang2012sparse}. 
These recurring patterns form a compact vocabulary of reusable \textbf{units}. At a longer time scale, units are organized into recurring \textbf{motifs}. Different instances of the same motif may contain repeated or missing units, or have slightly different boundaries, while still preserving a similar overall organization~\citep{kuehne2014language}.
This phenomenon appears across diverse domains, including human motion, driving, and logistics. 
For example, daily activities such as food preparation may reuse local motion units such as reaching for objects or changing posture, while differing in how these units are ordered and combined into longer motifs \citep{kipf2019compile, niemann2020lara,martin2019drive,chavarriaga2013opportunity,maes2021learning}.
We refer to this nested organization as \textbf{multiscale compositional recurrence}, illustrated in Fig.~\ref{fig:teaser}(a). Explicitly separating these levels allows shared local structure to be reused across multiple higher-level patterns rather than treating each realization as an independent discovery: accumulated data refine the unit vocabulary, while novel arrangements expand the motif vocabulary.

Existing unsupervised methods capture individual aspects of this structure but rarely model the interaction between temporal scales. Early clustering and discovery approaches focus on recurring local states within a sequence \citep{matsubara2014autoplait,hallac2017toeplitz}. Motif-discovery methods extend this perspective to repeated subsequences over longer horizons \citep{minnen2006discovering}, while more recent representation-learning and hierarchical segmentation approaches capture temporal structure at multiple resolutions or recover reusable components \citep{yue2022ts2vec,liu2024timesurl,kipf2019compile,wu2022learning}.
However, these levels are typically discovered or represented in isolation, leaving the connections between local structure and longer recurring patterns largely implicit. 
As illustrated in Fig.~\ref{fig:teaser}(b), this can fragment related temporal patterns into separate discoveries when the same local structures occur with different repetitions and arrangements.

In this work, we explicitly model the hierarchical organization of temporal streams. We introduce \projectname{}, an approach that discovers and maintains complementary vocabularies at different scales (Fig.~\ref{fig:teaser}(c)). At the local level, \textbf{UnitAlign} adapts generic temporal representations to learn a vocabulary of recurring and reusable units, while \textbf{MotifFormer} models their temporal organization to discover higher-level motifs. This separation enables local units to be reused across different motifs while preserving their distinct compositions.
As new data arrive, UnitAlign first updates the unit vocabulary by refining matched units or introducing recurring novel ones; MotifFormer then operates on the updated unit sequence to refine existing motifs or add new compositions.

We evaluate \projectname{} across diverse temporal datasets against unsupervised clustering, segmentation, and recurring-pattern discovery approaches. Across datasets, \projectname{} improves the recovery of recurring structure, with particularly strong gains when data are initially limited and subsequently expanded. Ablation studies show that both unit-level adaptation and higher-level motif modeling contribute to these improvements, while experiments with different temporal representations demonstrate that the gains are not tied to a particular representation backbone.

Our contributions are:
\begin{itemize}

\item We formulate unsupervised recurring-structure discovery as hierarchical temporal modeling across multiple scales with explicitly reusable local structure.

\item We introduce \projectname{}, combining \textbf{UnitAlign} for reusable unit discovery and \textbf{MotifFormer} for modeling their higher-level recurring compositions.

\item We demonstrate consistent improvements over existing unsupervised baselines across diverse temporal datasets, sensing modalities, and sequence characteristics.

\item We show that learned unit vocabularies can transfer across records to support motif discovery in new streams without relearning local structure.

\item We enable continual refinement and expansion of unit and motif vocabularies as new recurring structure emerges over evolving temporal streams.

\end{itemize}

%% file: sections/relatedwork.tex
\section{Related Work}
\label{sec:rw}

\subsection{Unsupervised Temporal Structure Discovery}
Recurring structure in temporal data has traditionally been studied through motif discovery. Classical methods identify approximately repeated subsequences and extend this formulation to multidimensional, variable-length, and streaming settings \citep{chiu2003probabilistic,mueen2009exact,mueen2010online,yeh2016matrix,linardi2018matrix}. 
These methods locate repeated temporal patterns directly from raw data without predefined semantic categories or boundaries, but generally treat motifs as independent subsequences. As a result, related structure that varies in duration, context, or composition may be fragmented across multiple discovered patterns.

More recent approaches incorporate richer temporal context when modeling long sequences. TICC, Time2State, and E2USD infer recurring states or regimes from multivariate streams, with E2USD further supporting efficient online state discovery \citep{hallac2017toeplitz,wang2023time2state,lai2024e2usd}. Unsupervised action segmentation combines representation learning, clustering, and temporal constraints to obtain coherent segment sequences \citep{kumar2022unsupervised,xu2024temporally}, while TNC, TS2Vec, and TimesURL learn contextual representations over different temporal supports \citep{tonekaboni2021unsupervised,yue2022ts2vec,liu2024timesurl}. 
Despite these advances, recurring structure is still typically represented as a flat set of states or segments, obscuring reuse across temporal scales. We instead explicitly model hierarchical relationships between recurring structures, allowing shared local patterns to be reused within longer temporal organizations.

\subsection{Hierarchical and Compositional Temporal Modeling}

Hierarchical representations have been widely used to model sequential data at multiple temporal scales. Rather than treating a long sequence as a collection of independent states, these approaches represent longer temporal patterns as compositions of shorter, reusable structures \citep{sutton1999between,niekum2012learning,kipf2019compile,jiang2022learning}. Such representations allow local structure to be shared across different temporal contexts and provide a natural way to capture compositional organization in long sequences.

This hierarchical view has also been explored in unsupervised temporal discovery. Existing approaches construct higher-level structure by recursively grouping lower-level temporal units, modeling compositional relations among discovered segments, or deriving coarse patterns from finer-grained structure \citep{wu2022learning,gong2023activity,spurio2025hierarchical}. Related work in smart-home sensing similarly reuses previously discovered local patterns to construct higher-level activity representations \citep{hiremath2022bootstrapping}. These studies show that discovered temporal structure need not be represented as a flat set of patterns. However, most existing approaches learn such hierarchies from fixed datasets or within specific domains, with limited support for revising them as new recurring structure emerges. We extend this direction to continual unsupervised discovery, where both local and higher-level structure can evolve with the continual stream.

\subsection{Continual and Transferable Structure Discovery}
Temporal discovery becomes more challenging when the stream arrives progressively. Classical stream-clustering methods such as CluStream and DenStream maintain and update cluster structure over evolving data streams \citep{aggarwal2003framework,cao2006density}, while online motif discovery and systems such as StreamScope incrementally track recurring temporal patterns \citep{mueen2010online,kawabata2018streamscope}. Related challenges have been studied in open-world and generalized category discovery, where models distinguish known from unseen categories and, in continual or online settings, incorporate novel categories while preserving previously discovered structure \citep{vaze2022generalized,wu2023metagcd,park2024online,ma2024happy,dai2025continual}.

Despite their different settings, these approaches generally model novelty at a single level, assigning each segment to an atomic category; recent temporal action-discovery methods similarly use labeled known actions to define the granularity of novel ones \citep{spurio2026looking}. This is restrictive for compositional streams, where novelty may arise within known motifs, through new compositions of known units, or as genuinely new units, while local structure may remain reusable even when higher-level compositions change. We therefore maintain separate unit- and motif-level vocabularies, allowing structure to be updated and transferred where recurring regularities persist or emerge.

%% file: sections/method.tex
\section{Methodology}
\label{sec:method}

\begin{figure}[t]
    \centering
    \includegraphics[width=\linewidth]{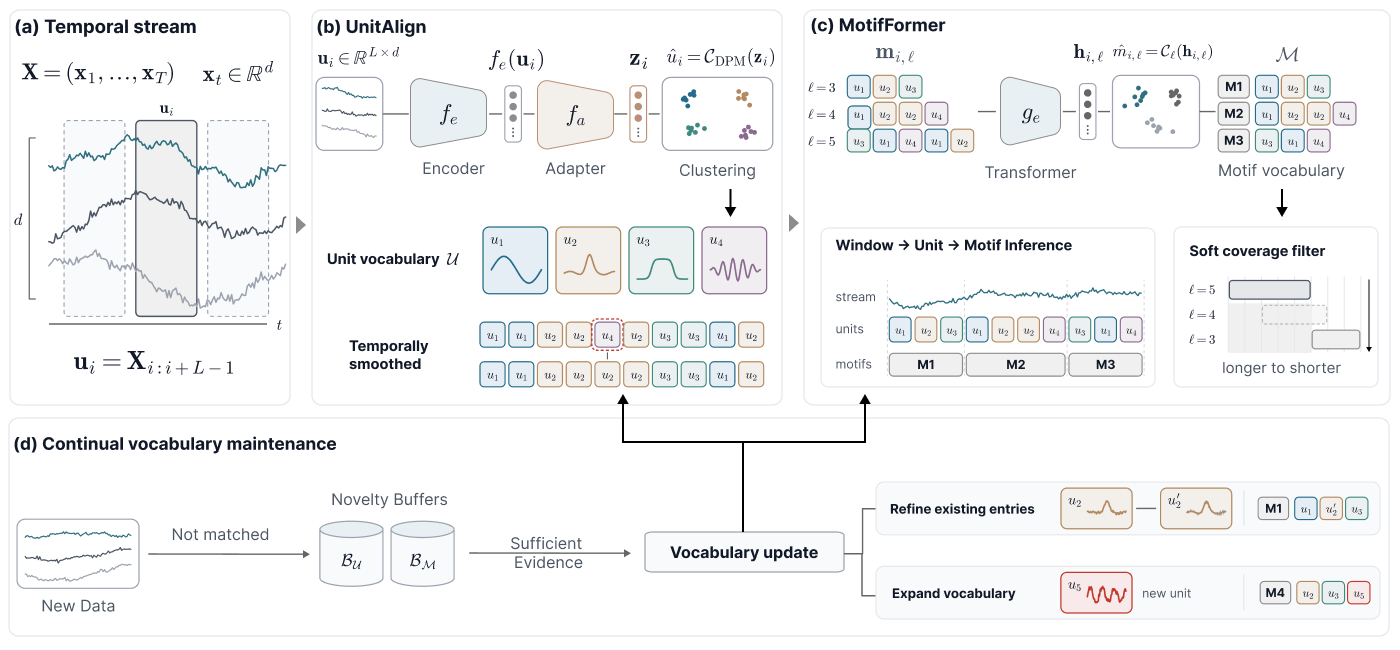}
    \caption{Overview of \projectname{}.
(a) A continuous temporal stream is divided into sliding windows. (b) UnitAlign encodes and adapts each window, clusters recurring local structure into a reusable unit vocabulary, and smooths unit assignments. (c) MotifFormer models unit sequences across temporal scales to discover higher-level motifs, followed by a soft coverage filter. (d) During continual operation, unmatched windows and unit subsequences are stored in novelty buffers and, once sufficient evidence is available, used to refine existing entries or introduce new units and motifs.}
    \label{fig:overview}
\end{figure}
We now describe \textbf{\projectname{}}, an unsupervised approach for discovering recurring, hierarchical temporal structure from continuous timeseries. As illustrated in Fig.~\ref{fig:overview}, the approach consists of two main stages. \textbf{UnitAlign} adapts local temporal representations and organizes recurring patterns into a shared unit vocabulary. \textbf{MotifFormer} then models the resulting unit sequence to discover recurring higher-level motifs. As the stream grows, both vocabularies are progressively updated. We first formalize the discovery problem before describing each component in detail.

\subsection{Problem Formulation}
Let a continuous multivariate temporal stream be
\begin{equation}
\mathbf{X}=(\mathbf{x}_1,\ldots,\mathbf{x}_T),
\qquad
\mathbf{x}_t\in\mathbb{R}^{d},
\label{eq:stream}
\end{equation}

where $T$ is the stream length and $d$ 
the number of channels. The general objective of unsupervised recurring pattern discovery is to identify temporal patterns that repeatedly occur in $\mathbf{X}$ without known temporal boundaries or semantic labels. We define each motif instance and its latent type as

\begin{equation}
\mathbf{m}_i
=
\mathbf{X}_{s_i:e_i},
\qquad
\hat{m}_i\in \mathcal{M}= \{1,\ldots,K_m\},
\label{eq:pattern}
\end{equation}

where $s_i$ and $e_i$ denote the unknown temporal boundaries of instance $\mathbf{m}_i$, and $\hat{m}_i$ denotes its motif class. 
Instances assigned to the same $\hat{m}$ share similar temporal structure despite variations in duration or boundaries. 
The task is to discover recurring patterns and their occurrences without knowing their number or temporal extent in advance. 
Temporal recurrence may occur across multiple scales, where shorter reusable structures are repeatedly composed into longer patterns of variable extent \citep{sanzari2019discovery, minnen2007discovering, li2012visualizing}. We refer to this nested structure as \textbf{multiscale compositional recurrence}.
To model this structure, we formulate recurring pattern discovery at two levels: reusable local \textbf{units} and higher-level \textbf{motifs}. As illustrated in Fig.~\ref{fig:overview}(a), we partition $\mathbf{X}$ into
sliding windows $\mathbf{u}_i$, which provide the basic inputs
for discovering reusable units.

\begin{equation}
\mathbf{u}_i
=
\mathbf{X}_{i:i+L-1},
\qquad
\hat{u}_i=f(\mathbf{u}_i),
\quad
\hat{u}_i\in\mathcal{U}=\{1,\ldots,K_u\}.,
\label{eq:unit}
\end{equation}

where \(L\) is the local window length, \(f(\cdot)\) maps each window to a unit class
and $\mathcal{U}$ is the discovered unit vocabulary. Applying Eq.~\ref{eq:unit} across the stream produces the ordered unit sequence, where recurring windows share entries in
the unit vocabulary $\mathcal{U}$.
At the higher level, each motif ($\hat{m}$) is represented by variable-length compositions of these units,
The recurring motif types $\hat{m}$ defined above are then represented at the motif level as variable-length compositions over $\mathcal{U}$. A motif instance is

\begin{equation}
\mathbf{m}_{i,l}
\mapsto
(\hat{u}_{i},\ldots,\hat{u}_{i+\ell-1}),
\qquad
\hat{m}_{i,\ell}=g(\mathbf{m}_{i,\ell}),
\qquad
\hat{m}_{i,\ell}\in\mathcal{M}
\label{eq:motif}
\end{equation}

where \(\ell\) is the number of constituent units, \(g(\cdot)\) maps each unit sequence to a motif.

\subsection{Discovering Reusable Temporal Units}

We first instantiate the unit mapping $f(\cdot)$ in Eq.~\ref{eq:unit} using
\textbf{UnitAlign}, illustrated in Fig.~\ref{fig:overview}(b).
Given each temporal window $\mathbf{u}_i$ from Fig.~\ref{fig:overview}(a), a pretrained self-supervised encoder $f_e$ extracts an initial representation,
which an adaptor $f_a$ maps into the latent embedding $\mathbf{z}_i$. DeepDPM then clusters these embeddings and assigns each window to a unit:
\begin{equation}
\mathbf{z}_i=f_a\!\left(f_e(\mathbf{u}_i)\right),
\qquad
\hat{u}_i=\mathcal{C}_{\mathrm{DPM}}(\mathbf{z}_i),
\quad \hat{u}_i\in\mathcal{U},
\label{eq:unit_embedding}
\end{equation}
where $\mathbf{z}_i$ is the adapted representation and
$\mathcal{C}_{\mathrm{DPM}}$ denotes the DeepDPM cluster assignment.
Each resulting cluster corresponds to one entry in the unit vocabulary
$\mathcal{U}$ shown in Fig.~\ref{fig:overview}(b).

The adaptor is optimized with two complementary objectives. A consistency objective preserves the similarity structure of the original SSL representations, while DeepDPM provides pseudo-labels that encourage separability in the adapted space. We alternate between representation adaptation and clustering until the assignments stabilize.
Finally, we temporally smooth the resulting assignments to suppress isolated
transitions, producing the stable unit sequence at the bottom of
Fig.~\ref{fig:overview}(b), which serves as input to MotifFormer.

\subsection{Discovering Compositional Motifs}
We then instantiate the motif mapping $g(\cdot)$ in Eq.~\ref{eq:motif} using
\textbf{MotifFormer}, illustrated in Fig.~\ref{fig:overview}(c).
Starting from the smoothed unit sequence produced by UnitAlign, we construct
unit subsequences of different lengths $\ell$. A Transformer encoder $g_e$
maps each subsequence $\mathbf{m}_{i,\ell}$ into a motif representation
$\mathbf{h}_{i,\ell}$, which is then clustered into a recurring motif:

\begin{equation}
\mathbf{h}_{i,\ell}
= g_e(\mathbf{m}_{i,l})=
g_e(\hat{u}_i,\ldots,\hat{u}_{i+\ell-1}),
\qquad
\hat{m}_{i,\ell}
=
\mathcal{C}_{\ell}(\mathbf{h}_{i,\ell}),
\quad
\hat{m}_{i,\ell}\in\mathcal{M},
\label{eq:motif_embedding}
\end{equation}
where $\mathcal{C}_{\ell}$ denotes the clustering assignment at temporal scale $\ell$. The Transformer captures contextual dependencies among units while mapping variable-length sequences into a shared space.

To make these representations robust across local variation and temporal
extent, MotifFormer is optimized with two complementary objectives. A consistency objective encourages representations to remain stable under perturbations of the unit sequence and aligns subsequences anchored at the same starting point across different lengths, promoting continuity across temporal scales. In parallel, clustering-derived pseudo-labels are used to improve the separability of motif representations.

After training, we cluster subsequences separately at each temporal scale.
As shown in the lower-right of Fig.~\ref{fig:overview}(c), a
\textbf{soft coverage filter} processes candidates from longer to shorter
scales, downweighting regions already explained by high-confidence long motifs.
This reduces redundant rediscovery of already captured structure and allows shorter recurring patterns to emerge. 
Finally, we aggregate evidence across scales based on each subsequence's proximity to its cluster center to determine the final motif boundaries and assignments.

\subsection{Continual Vocabulary Maintenance}

Finally, Fig.~\ref{fig:overview}(d) shows how both vocabularies evolve as the stream evolves. When a new window or unit subsequence cannot be confidently matched to an existing entry, it is stored in separate unit and motif novelty buffers, $\mathcal{B}_{U}$ and $\mathcal{B}_{M}$, rather than immediately creating a new vocabulary item.

We manage this evolution through dynamically maintained micro-clusters that summarize and adapt the current vocabulary structure. Once sufficient evidence accumulates, buffered windows and subsequences are integrated with the existing micro-clusters to trigger a vocabulary update. The update either \emph{refines} an existing entry when the buffered data remain compatible with known structure, or \emph{expands} the vocabulary when a persistent new unit or motif emerges.

%% file: sections/experiment.tex
\section{Experiments}
\label{sec:experiments}
Our experiments address three questions: (1) how accurately can \projectname{} recover recurring temporal structure compared with existing unsupervised methods; (2) how effectively can it adapt as new data emerge; and (3) which components and design choices contribute most to its performance.

\subsection{Experimental Setup}
\label{sec:setup}

\textbf{Datasets.}
We evaluate \projectname{} on six multivariate temporal 
datasets: \textbf{MoCap}, \textbf{ActRecTut}, \textbf{PAMAP2}, \textbf{Drive\&Act}, \textbf{OPPORTUNITY}, and \textbf{LARa}~\citep{zhou2012hierarchical,bulling2014tutorial,reiss2012introducing,martin2019drive,chavarriaga2013opportunity,niemann2020lara,rueda2021human}. These datasets span diverse sensing environments, class vocabularies, and sequence characteristics, allowing us to evaluate recurring structure discovery across heterogeneous continuous streams. We treat individual recordings as independent sequences and apply dataset-specific preprocessing and window configurations summarized in Appendix~\ref{app:data}.

\textbf{Baselines.} We compare \projectname{} against unsupervised temporal state and structure discovery methods, including \textbf{AutoPlait}~\citep{matsubara2014autoplait}, \textbf{TICC}~\citep{hallac2017toeplitz}, \textbf{Time2State}~\citep{wang2023time2state}, \textbf{E2USD}~\citep{lai2024e2usd}, \textbf{CompILE}~\citep{kipf2019compile}, and \textbf{CLaP}~\citep{ermshaus2025clap}. For continual discovery, we additionally compare against the streaming clustering methods \textbf{DenStream}~\citep{cao2006density} and \textbf{DBSTREAM}~\citep{hahsler2016clustering}. Together, these baselines cover regime segmentation, representation-based state discovery, temporal discovery, and online density-based clustering.

\textbf{Implementation Details.}
We use the publicly released implementations of the baselines and follow their original configurations whenever applicable. To ensure a consistent comparison, all methods operate on the same recordings and are evaluated on a common temporal axis. We retain each method's original learning and inference procedure while adapting input preprocessing and dataset-specific configurations where necessary. Detailed preprocessing choices, hyperparameters, and implementation settings are provided in Appendices~\ref{app:data} and \ref{app:baselines}.

\subsection{Evaluation Protocol and Metrics}
\label{sec:metrics}
We evaluate all methods on a common temporal axis using a per-recording many-to-one mapping from discovered clusters to ground-truth classes. Each cluster is assigned to the class with the largest duration-weighted temporal overlap, allowing multiple clusters to represent different realizations of the same underlying pattern. This mapping is used only for evaluation and follows prior over-clustering and temporal-discovery protocols \citep{peng2020mutual,chen2025unsupervised}. The \textbf{Other} label is retained as a background class.

We report MoF and discovery recall for overall discovery quality, and Edit and segmental F1 at IoU thresholds of 10\%, 25\%, and 50\% for temporal consistency \citep{lea2016segmental,lea2017temporal}. We additionally report the segment ratio (SegRatio), defined as the number of predicted segments relative to ground-truth segments, to characterize over- and under-segmentation \citep{krishna2014temporal,furnari2017shall}. For continual discovery, we report H-score and forgetting $M_f$ to measure adaptation to emerging structure and retention of previously discovered patterns \citep{vaze2022generalized,wen2023parametric,wu2023metagcd,park2024online}. The main text reports MoF, Recall, Edit, F1@50, and SegRatio; complete results are provided in the Appendix~\ref{tab:full}.

\subsection{Recurring Structure Discovery}
\label{sec:main_results}
We first evaluate whether \projectname{} can recover recurring structure from temporal streams.
We compare against six baselines across all six datasets using the evaluation protocol described in Section~\ref{sec:metrics}. This experiment evaluates both discovery quality and temporal segmentation, capturing whether recurring structure is recovered with coherent organization.

As shown in Table~\ref{tab:main_results}, \textsc{Rondo} achieves the highest Recall on all six datasets and the highest MoF on five, slightly trailing only AutoPlait on MoCap, suggesting that it recovers recurring structure more completely and assigns it consistently over time.
This advantage holds across datasets with diverse modalities and temporal characteristics. Segment-level results show a more mixed pattern. SegRatio indicates that AutoPlait and CLaP often produce fewer, longer segments, while Time2State and E2USD tend to split the sequence into many short segments. \projectname{} lies between these two extremes and achieves strong F1 and Edit scores on several datasets. PAMAP2 is an exception, where \projectname{} recovers the underlying classes well but has a high SegRatio, indicating over-segmentation. Overall, the results suggest that modeling recurring structure at multiple levels improves discovery while maintaining reasonably coherent temporal segmentation.

\begin{table}[t]
\centering
\caption{
Recurring structure discovery results (\%). Stochastic methods report mean performance over five random seeds, with standard deviations provided in Appendix~\ref{app:seeds}; deterministic or officially fixed-seed methods are evaluated once. CompILE is evaluated once due to its computational cost. Higher is better for MoF, Recall, F1@50, and Edit. For SegRatio, values closer to 1 are better, so ranked with $|\log \mathrm{SegRatio}|$ and lower is better. Best in \textbf{bold}, second-best underlined.
}
\label{tab:main_results}
\scriptsize

\begin{tabular}{lccccc @{\hspace{8pt}} lccccc}
\toprule
Method & MoF & Recall & F1@50 & Edit & SegRatio & Method & MoF & Recall & F1@50 & Edit & SegRatio \\
\midrule
\multicolumn{6}{c}{\textbf{MoCap}} &
\multicolumn{6}{c}{\textbf{Drive\&Act}} \\
\cmidrule(r){1-6}\cmidrule(l){7-12}
Time2State & 88.68 & \underline{91.49} & 84.48 & 83.80 & 1.70 & 
Time2State & \underline{54.23} & \underline{29.55} & 7.00 & 12.97 & 11.31 \\ 
E2USD & 84.97 & 87.45 & 77.95 & 76.48 & 1.97 &
E2USD & 47.60 & {18.90} & 5.66 & 11.80 & 14.47 \\ 
TICC & 86.61 & 84.52 & 79.39 & 83.18 & 2.19 & 
TICC & 49.47 & 23.71 & 6.61 & {13.01} & 9.14 \\ 
AutoPlait & \textbf{90.27} & 90.28 & \underline{91.18} & \underline{92.05} & \textbf{0.97}& 
AutoPlait & 39.42 & 8.08 & 0.26 & 4.32 & 0.04\\ 
CompILE & 66.45 & 61.05 & 52.19 & 63.14 & 1.97 & 
CompILE & 46.17 & 17.20 & 9.04 & \textbf{32.69} & \underline{4.13} \\  
CLaP & 74.61 & 65.63 & 60.53 & 62.39 & \underline{0.74}& 
CLaP & 49.26 & 17.74 & \underline{11.43} & 20.10 & \textbf{0.37} \\ 
\projectname{} & \underline{88.91} & \textbf{96.78} & \textbf{95.06} & \textbf{96.09} & 2.42& 
\projectname{} & \textbf{68.36} & \textbf{62.30} & \textbf{14.29} & \underline{22.22} & 8.04 \\ 
\midrule
\multicolumn{6}{c}{\textbf{PAMAP2}}&
\multicolumn{6}{c}{\textbf{OPPORTUNITY}} \\
\cmidrule(r){1-6}\cmidrule(l){7-12}
Time2State & \underline{79.69} & \underline{92.13} & 6.17 & 8.65 & 24.72 &
Time2State & \underline{69.28} & {48.94} & {20.80} & \textbf{58.15} & \underline{2.08}\\ 
E2USD & 76.58 & 90.21 & 4.22 & 7.25 & 26.96 & 
E2USD & {69.02} & \underline{49.95} & \underline{26.49} & \underline{50.32} & 2.91 \\ 
TICC & 74.71 & 78.69 & 15.93 & 16.42 & 7.27 & 
TICC & 65.91 & 42.99 & 13.66 & 21.28& 0.37 \\ 
AutoPlait & 45.22 & 29.81 & {21.46} & \underline{21.45} & \textbf{0.97}&
AutoPlait & 37.64 & 20.83 & 0.00 & 0.63 & 0.01 \\ 
CompILE & 34.11 & 8.28 & 4.65 & 10.41 & 15.92 & 
CompILE & 47.13 & 25.54 & 3.44 & 25.24 & 0.46 \\  
CLaP & 77.40 & 79.70 & \textbf{50.00} & \textbf{51.30} & \underline{3.03}& 
CLaP & 58.88 & 29.61 & 2.82 & 6.61 &0.09\\ 
\projectname{} & \textbf{84.12} & \textbf{94.25} & \underline{22.91} & {20.82} & 15.37& 
\projectname{} & \textbf{79.96} & \textbf{59.19} & \textbf{30.59} & 48.06& \textbf{1.23} \\ 
\midrule
\multicolumn{6}{c}{\textbf{ActRecTut}}&
\multicolumn{6}{c}{\textbf{LARa}} \\
\cmidrule(r){1-6}\cmidrule(l){7-12}
Time2State & \underline{73.40} & \underline{49.08} & \textbf{42.20} & \textbf{49.07} & \underline{1.10} & 
Time2State & {69.91} & \underline{37.95} & {27.09} & \underline{45.89} & 2.00 \\ 
E2USD & 72.87 & 46.83 & \underline{40.41} & 45.03 & \textbf{1.01} & 
E2USD & {67.24} & {35.62} & 21.81 & 40.36 & 2.13\\ 
TICC & 69.68 & 41.59 & 29.33 & 32.24 & 0.45&
TICC & \underline{70.40} & 35.57 & \underline{28.34} & {44.85}&1.92 \\ 
AutoPlait & 44.42 & 14.06 & 2.04 & 2.31 & 0.01&
AutoPlait & 57.46 & 17.22 & 1.36 & 5.06&0.04 \\ 
CompILE & 45.69 & 27.96 & 8.76 & 31.11 & 0.44 & 
CompILE & 60.03 & 31.30 & 11.20 & 26.13 & \underline{0.57} \\  
CLaP & 51.68 & 27.70 & 5.45 & 5.45 &0.02& 
CLaP & 65.28 & 23.93 & 7.47 & 10.76&0.16 \\ 
\projectname{} & \textbf{75.66} & \textbf{52.85} & 36.71 & \underline{45.26} &0.77& 
\projectname{} & \textbf{78.44} & \textbf{53.01} & \textbf{38.50} & \textbf{47.34} & \textbf{1.31}\\ 
\bottomrule
\end{tabular}
\end{table}

\subsection{Adaptation and Transfer across Streams}
\label{sec:continual}

\textbf{Continual discovery under progressive streams.}
We evaluate whether \projectname{} can preserve and extend previously discovered structure as new observations become available. We focus on \textbf{OPPORTUNITY} and \textbf{PAMAP2}, which provide long continuous recordings suitable for progressive evaluation. For each dataset, only the first 20\% of the stream is used to initialize the unit and motif vocabularies, after which the remaining observations are revealed incrementally. Since representations learned independently from each complete recording, such as TS2Vec embeddings, would expose the model to future observations, we instead use the pretrained MantisV2 encoder in this experiment (Appendix~\ref{app:encoders}).

We compare against the streaming clustering methods \textbf{DenStream} and \textbf{DBSTREAM} under the same progressive schedule, together with \textbf{Time2State@20} as a static reference trained only on the initial portion of the stream. In addition to standard discovery metrics, we report H-score throughout the stream to characterize continual adaptation and $M_f$ to measure forgetting. As shown in Table~\ref{tab:continual_results}, \projectname{} remains competitive in frame-level discovery while producing substantially more coherent temporal structure, with stronger segment-level quality and considerably less fragmentation than the streaming baselines.

This advantage becomes more apparent as additional observations accumulate: \projectname{} achieves stronger H-scores at later stages of the stream, suggesting that its unit and motif vocabularies can be progressively refined and expanded while retaining previously discovered structure. These results highlight the benefit of explicitly maintaining reusable temporal structure as the stream evolves, rather than treating continual discovery solely as online clustering.

\begin{table}[t]
\centering
\caption{Continual discovery on progressively revealed streams. H@40--100 denotes H-score after observing 40--100\% of the stream.}
\label{tab:continual_results}
\scriptsize
\begin{tabular}{lccccc @{\hspace{8pt}} ccccc}
\toprule
Method & MoF & Recall & F1@50 & Edit & SegRatio & H@40 & H@60 & H@80 & H@100 & $M_f$ \\
\midrule
\multicolumn{11}{c}{\textbf{OPPORTUNITY}} \\
\midrule
Time2State@20 & 44.70 & 36.20 & 7.00 & \underline{36.00} & \underline{5.58} & 0.00 & 0.00 & 32.00 & 20.50 & 24.00 \\
DBSTREAM@20 & 47.10 & 34.00 & 8.00 & 26.20 & 6.02 & \textbf{30.50} & 36.60 & 44.60 & 39.00 & 25.80 \\
DenStream@20 & \textbf{59.90} & \textbf{45.90} & \underline{15.10} & 23.56 & 8.75 & 11.40 & \underline{37.60} & \underline{57.80} & \underline{58.30} & \textbf{12.40} \\
\projectname{}@20 & \underline{58.10} & \underline{42.90} & \textbf{21.10} & \textbf{44.70} & \textbf{4.03} & \underline{23.40} & \textbf{38.70} & \textbf{65.40} & \textbf{63.50} & \underline{14.60} \\
\midrule
\multicolumn{11}{c}{\textbf{PAMAP2}} \\
\midrule
Time2State@20 & 41.10 & 23.70 & 1.10 & \textbf{6.14} & \textbf{27.71} & 50.10 & 40.30 & 34.90 & 32.60 & \underline{6.20} \\
DBSTREAM@20 & 55.00 & 52.10 & \textbf{3.30} & \underline{4.71} & \underline{38.57} & 59.60 & 52.90 & 49.50 & 47.00 & 19.70 \\
DenStream@20 & \underline{61.90} & \underline{63.80} & \underline{1.90} & 3.87 & 46.73 & \underline{64.50} & \underline{62.90} & \underline{61.30} & \underline{53.30} & 14.60 \\
\projectname{}@20 & \textbf{69.04} & \textbf{75.20} & 1.17 & 3.44 & 59.57 & \textbf{73.70} & \textbf{72.49} & \textbf{71.59} & \textbf{70.51} & \textbf{4.69} \\
\bottomrule
\end{tabular}
\end{table}

\textbf{Cross-record transfer of unit vocabularies.}
We further examine whether the discovered local units capture reusable temporal structure beyond the stream from which they are learned. We conduct this experiment on \textbf{LARa} and \textbf{Drive\&Act}, both of which contain recordings from multiple participants and therefore provide a natural setting for evaluating transfer under changes in motif execution. 
Table~\ref{tab:cu_results} shows the cross-record transfer results. In the standard transfer setting, all methods learn from a source participant and directly apply the discovered structure to unseen target participants, a limited-data setting that tests how well their discovered patterns generalize across records.
Under this setting, \projectname{} outperforms Time2State and E2USD on both datasets, suggesting that its discovered structure generalizes effectively across records. More importantly, our hierarchical formulation allows us to transfer only the learned unit vocabulary and discover new motifs from the target stream. This \projectname{}-fit variant further improves MoF and Recall on both datasets, indicating that local units remain reusable across records while higher-level motifs can be adapted to user-specific temporal compositions. This separation provides a path toward personalizing temporal structure discovery from limited target data without relearning the entire representation from scratch.

\begin{table}[t]
\centering
\caption{Cross-record transfer results on LARa and Drive\&Act. \projectname{}-fit updates only MotifFormer on the target record.}
\label{tab:cu_results}
\scriptsize
\begin{tabular}{lccccc @{\hspace{8pt}} lccccc}
\toprule
Method & MoF & Recall & F1@50 & Edit & SegRatio &
Method & MoF & Recall & F1@50 & Edit & SegRatio \\
\midrule
\multicolumn{6}{c}{\textbf{LARa}} &
\multicolumn{6}{c}{\textbf{Drive\&Act}} \\
\cmidrule(r){1-6}\cmidrule(l){7-12}

Time2State
& 50.62 & 28.56 & 16.67 & 40.72 & 1.59
&
Time2State
& 44.80 & 17.53 & 4.99 & 14.07 & 14.37 \\

E2USD
& 45.59 & 21.20 & 12.06 & 41.64 & 1.78
&
E2USD
& 43.75 & 14.94 & 6.00 & 13.34 & 14.68 \\

\projectname{}
& \underline{62.20} & \underline{43.89} & \underline{29.72} & \underline{45.32} & \textbf{1.24}
&
\projectname{}
& \underline{54.27} & \underline{38.11} & \underline{8.97} & {\textbf{23.42}} & \textbf{7.98} \\
\cmidrule(r){1-6}\cmidrule(l){7-12}
\projectname{}-fit
& {\textbf{63.48}} & {\textbf{44.05}} & \textbf{31.35} & {\textbf{47.83}} & \underline{1.25}
&
\projectname{}-fit
& {\textbf{58.66}} & {\textbf{46.50}} & {\textbf{9.84}} & \underline{18.94} & \underline{9.28} \\

\bottomrule
\end{tabular}
\end{table}

\subsection{Understanding Rondo}
\label{sec:analysis}

We further analyze \projectname{} from two perspectives: the contribution of its two main components and the effect of the underlying temporal representation.

\paragraph{Component ablation.}
We first evaluate the contribution of \textbf{UnitAlign} and \textbf{MotifFormer} by removing each component separately from the full approach. Removing UnitAlign tests the importance of adapting local representations to the recurring structure of the target stream, while removing MotifFormer eliminates explicit modeling of higher-level compositions. As shown in Table~\ref{tab:ablation}, the full model achieves the highest MoF across all four datasets, indicating that both components contribute to more consistent structure discovery. Removing MotifFormer degrades MoF across datasets and leads to substantial drops in segment-level performance on PAMAP2 and Drive\&Act, highlighting the benefit of explicitly modeling compositions of reusable units. Removing UnitAlign particularly reduces discovery quality on LARa and Drive\&Act, while the less uniform effects on F1@50 and Edit suggest that representation adaptation primarily improves recurring-structure recovery rather than boundary estimation.

\begin{table}[t]
\centering
\caption{Ablation of \projectname{}.
\textbf{w/o UA} removes UnitAlign while retaining MotifFormer, and
\textbf{w/o MF} removes MotifFormer while retaining unit-level discovery.}
\label{tab:ablation}
\scriptsize

\begin{tabular}{lccccc @{\hspace{8pt}} lccccc}
\toprule
Method & MoF & Recall & F1@50 & Edit & SegRatio &
Method & MoF & Recall & F1@50 & Edit & SegRatio \\
\midrule

\multicolumn{6}{c}{\textbf{OPPORTUNITY}} &
\multicolumn{6}{c}{\textbf{LARa}} \\
\cmidrule(r){1-6}\cmidrule(l){7-12}

w/o MF
& \underline{78.70} & 58.20 & 28.90 & \textbf{49.48} & \textbf{1.13} &
w/o MF
& 76.60 & \underline{48.80} & 36.30 & \textbf{47.66} & 1.40 \\

w/o UA
& 76.90 & \textbf{65.40} & \textbf{35.00} & 47.11 & \underline{0.86} &
w/o UA
& \underline{77.60} & 45.60 & \underline{37.30} & 44.20 & {\textbf{1.26}} \\

\projectname{}
& \textbf{79.96} & \underline{59.19} & \underline{30.59}
& \underline{48.06} & {1.23} &
\projectname{}
& \textbf{78.44} & \textbf{53.01} & \textbf{38.50}
& \underline{47.34} & \underline{1.31} \\

\midrule

\multicolumn{6}{c}{\textbf{PAMAP2}} &
\multicolumn{6}{c}{\textbf{Drive\&Act}} \\
\cmidrule(r){1-6}\cmidrule(l){7-12}

w/o MF
& 80.30 & 92.60 & 11.12 & 12.94 & 17.99 &
w/o MF
& 55.90 & 35.99 & 7.66 & 13.12 & 11.69 \\

w/o UA
& \underline{82.43} & \textbf{94.36} & \underline{22.31}
& \textbf{21.82} & {\textbf{10.85}} &
w/o UA
& \underline{66.28} & \underline{55.84} & \textbf{15.72}
& \textbf{24.21} & {\textbf{7.45}} \\

\projectname{}
& \textbf{84.12} & \underline{94.25} & \textbf{22.91}
& \underline{20.82} & \underline{15.37} &
\projectname{}
& \textbf{68.36} & \textbf{62.30} & \underline{14.29}
& \underline{22.22} & \underline{8.04} \\

\bottomrule
\end{tabular}
\end{table}

\paragraph{Representation backbone.}
We next examine the sensitivity of \projectname{} to the choice of temporal representation encoder. We test the encoder within UnitAlign to TS2Vec~\citep{yue2022ts2vec}, Time-MAE~\citep{cheng2026timemae}, and MantisV2~\cite{feofanov2026mantisv2} while keeping the remaining discovery pipeline unchanged. As shown in Table~\ref{tab:representation}, performance remains generally stable across representation choices, although the strongest encoder varies by dataset. TS2Vec achieves the highest MoF on OPPORTUNITY and PAMAP2, while Time-MAE performs best on Drive\&Act and LARa and improves several segment-level metrics. MantisV2 is competitive on some segmentation metrics but shows less consistent discovery performance overall. These results suggest that \projectname{} is not tied to a particular representation model, while the quality of the underlying representation still influences the discovered temporal structure.

\begin{table}[t]
\centering
\caption{
Representation sensitivity of \projectname{} across four datasets. Results compare TS2Vec, Time-MAE, and MantisV2 under the same discovery pipeline, showing that performance remains competitive across different self-supervised and pretrained temporal representations.
}
\label{tab:representation}
\scriptsize

\begin{tabular}{lccccc @{\hspace{8pt}} lccccc}
\toprule
Method & MoF & Recall & F1@50 & Edit & SegRatio &
Method & MoF & Recall & F1@50 & Edit & SegRatio \\
\midrule

\multicolumn{6}{c}{\textbf{OPPORTUNITY}} &
\multicolumn{6}{c}{\textbf{LARa}} \\
\cmidrule(r){1-6}\cmidrule(l){7-12}

TS2Vec
& \textbf{79.96}
& 59.19
& \underline{30.59}
& \underline{48.06}
& \textbf{1.23}
&
TS2Vec
& \underline{78.44}
& \underline{53.01}
& 38.50
& 47.34
& \textbf{1.31}
\\

Time-MAE
& \underline{78.50}
& \textbf{65.30}
& 28.60
& 46.70
& \underline{1.36}
&
Time-MAE
& \textbf{79.10}
& \textbf{53.70}
& \textbf{46.10}
& \underline{57.30}
& \underline{2.14}
\\

MantisV2
& 72.30
& \underline{61.00}
& \textbf{31.30}
& \textbf{63.80}
& 2.40
&
MantisV2
& 74.60
& 51.50
& \underline{39.90}
& \textbf{58.90}
& 2.23
\\

\midrule

\multicolumn{6}{c}{\textbf{PAMAP2}} &
\multicolumn{6}{c}{\textbf{Drive\&Act}} \\
\cmidrule(r){1-6}\cmidrule(l){7-12}

TS2Vec
& \textbf{84.12}
& 94.25
& \textbf{22.91}
& \textbf{20.82}
& \textbf{15.37}
&
TS2Vec
& \underline{68.36}
& \underline{62.30}
& \underline{14.29}
& \textbf{22.22}
& \textbf{8.04}
\\

Time-MAE
& \underline{83.50}
& \underline{95.30}
& \underline{16.10}
& \underline{16.30}
& \underline{17.40}
&
Time-MAE
& \textbf{75.10}
& \textbf{68.10}
& \textbf{15.40}
& \underline{20.90}
& \underline{9.41}
\\

MantisV2
& 82.70
& \textbf{97.30}
& 15.20
& 14.90
& 21.56
&
MantisV2
& 64.80
& 61.30
& 7.30
& 13.70
& 12.92
\\

\bottomrule
\end{tabular}
\end{table}

%% file: sections/app_qual.tex
\section{Qualitative Analysis}
\label{app:qual}

\begin{figure}[H]
\centering
\includegraphics[width=\linewidth]{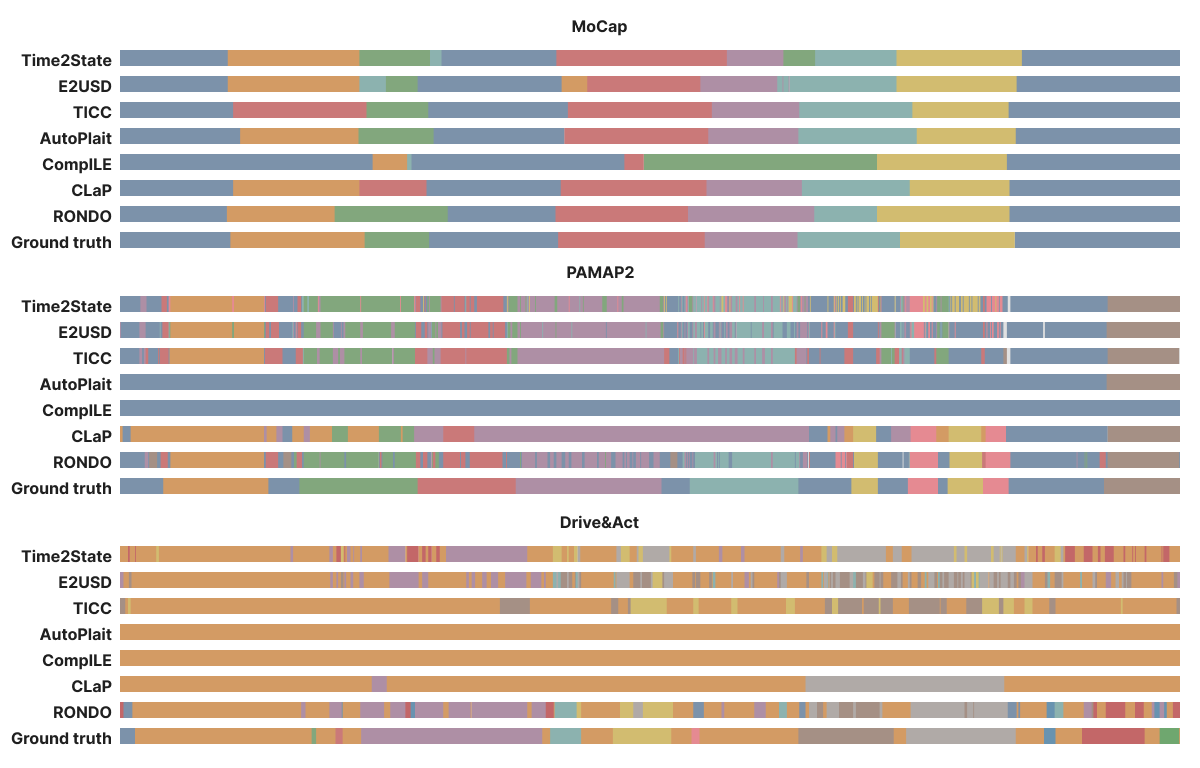}
\caption{Segmentation of one recording from three representative datasets, against the
ground truth (bottom). Predicted states are coloured by the class they map to;
a single-colour row means every predicted state mapped to the same class.}
\label{fig:segment}
\end{figure}

Beyond the quantitative metrics in the main text, we randomly sample recordings for qualitative inspection. Figure~\ref{fig:segment} shows that baseline discovery often struggles to balance local variation and long-range consistency: overly sensitive state changes lead to fragmented segments, whereas aggressive grouping merges distinct activities or collapses large portions of a sequence into a few states. In contrast, \projectname{} generally preserves clearer temporal structure while still capturing short activity transitions, although some long activities remain split into multiple recurring units.

We further inspect the relationship between the two hierarchical levels through their activity purity. Across most recordings, discovered units are temporally persistent, with around 75\% to over 90\% of adjacent windows retaining the same unit assignment. The unit vocabulary is also repeatedly reused within each recording, and in several datasets individual units already show relatively strong alignment with activity classes. This suggests that recurring local structure alone can often provide meaningful discrimination, particularly when activities exhibit distinctive and stable local dynamics.

The motif layer becomes more useful when this local alignment is weaker. On datasets where units are shared across several activities, motif-level grouping can improve activity consistency by roughly a few to more than ten percentage points, indicating that longer temporal context helps disambiguate locally similar patterns. In contrast, when unit-level purity is already high, motif composition provides only marginal gains, especially for short actions that are shorter than the modeled motif scale. These observations suggest that the hierarchy is not simply adding another clustering layer: units capture reusable local primitives, while motifs provide contextual disambiguation mainly when local patterns alone are insufficient.

%% file: sections/app_metrics.tex
\section{Evaluation Metrics}
\label{app:metrics}
For completeness, we provide the definitions of all evaluation metrics used in the paper. These metrics evaluate discovery quality, temporal segmentation, and continual adaptation from complementary perspectives.
\subsection{Temporal Alignment and Label Mapping}

\begin{figure}[h]
    \centering
    \includegraphics[width=\linewidth]{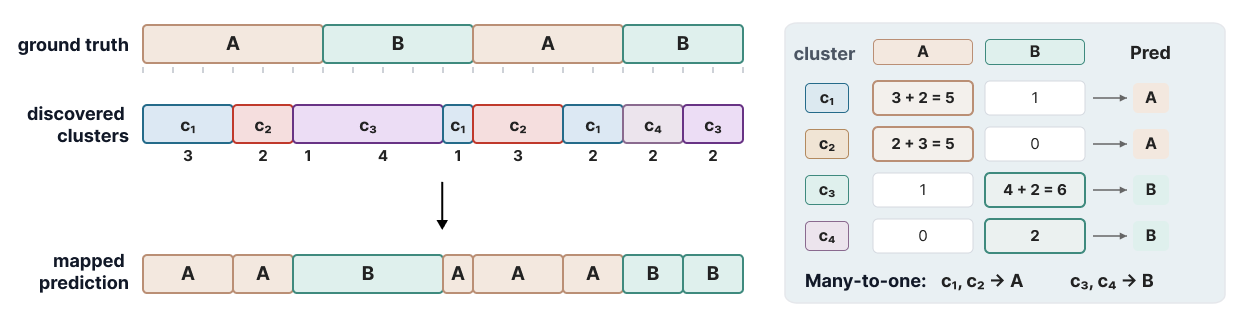}
    \caption{ Illustration of the evaluation-only many-to-one mapping. Discovered clusters
are matched to ground-truth labels by their duration-weighted overlap on a
common temporal axis, allowing multiple clusters to map to the same semantic
label.}
    \label{fig:mapping}
\end{figure}

All predictions and annotations are projected onto a common temporal axis consisting of consecutive intervals. Let $y_t$ denote the ground-truth class label and $\hat{z}_t$ the discovered cluster at interval $t$. Ground-truth labels are used only for evaluation and are not involved in training, clustering, or online updating.
Because different execution-specific compositions may correspond to the same semantic class, we use a many-to-one mapping from discovered clusters to ground-truth class labels, as illustrated in
Fig.~\ref{fig:mapping}. For each discovered cluster $c$, we assign the label with the largest duration-weighted overlap:

\begin{equation}
\pi(c) = \arg\max_{y}
\sum_{t}
\Delta_t
\mathbb{I}[\hat{z}_t=c \land y_t=y],
\end{equation}

where $\Delta_t$ is the duration of interval $t$. The mapped prediction is $\tilde{y}_t=\pi(\hat{z}_t)$. The mapping is estimated separately for each recording on the evaluation set and fixed across all continual stages. This evaluation-only many-to-one protocol follows prior over-clustering and action-cluster evaluation settings \citep{peng2020mutual,chen2025unsupervised}.

\subsection{Class and Segment Metrics}

We report duration-weighted mean over frame (MoF), discovery recall, Edit score, and segmental F1. MoF is computed as

\begin{equation}
\mathrm{MoF}=\frac{
\sum_{t\in\mathcal{T}*{\mathrm{eval}}}
\Delta_t,
\mathbb{I}[\tilde{y}*t=y_t]
}{
\sum*{t\in\mathcal{T}*{\mathrm{eval}}}
\Delta_t
}.
\end{equation}

Discovery recall measures the proportion of ground-truth class instances that are successfully discovered:

\begin{equation}
\mathrm{Recall}=\frac{\mathrm{TP}}
{\mathrm{TP}+\mathrm{FN}}.
\end{equation}

For the Edit score, consecutive intervals with the same mapped label are first merged into segments. Let $S$ and $\hat{S}$ be the ground-truth and predicted segment-label sequences. We compute

\begin{equation}
\mathrm{Edit}=100\left(
1-
\frac{
d_{\mathrm{Lev}}(\hat{S},S)
}{
\max(|\hat{S}|,|S|)
}
\right),
\end{equation}

where $d_{\mathrm{Lev}}$ is the Levenshtein distance \citep{lea2016segmental}.

For segmental F1, a predicted segment $p$ matches a ground-truth segment $g$ when they have the same mapped label and

\begin{equation}
\mathrm{IoU}(p,g)=\frac{|p\cap g|}{|p\cup g|}
\geq \tau.
\end{equation}

We report

\begin{equation}
\mathrm{F1}@\tau=\frac{2P_\tau R_\tau}{P_\tau+R_\tau},
\qquad
\tau\in{0.10,0.25,0.50},
\end{equation}

following standard temporal action segmentation practice \citep{lea2017temporal}. Segment matching is one-to-one, so duplicate predictions do not receive repeated credit.

\subsection{Continual Discovery Metrics}

We evaluate the model after each update stage using performance on previously known classes, newly discovered classes, and all classes. Let $A_{\mathrm{old}}^{(s)}$ and $A_{\mathrm{new}}^{(s)}$ denote the corresponding class-level performance at stage $s$.

The H-score measures the balance between retaining old classes and discovering new ones:

\begin{equation}
H^{(s)}=\frac{
2A_{\mathrm{old}}^{(s)}A_{\mathrm{new}}^{(s)}
}{
A_{\mathrm{old}}^{(s)}+A_{\mathrm{new}}^{(s)}
}.
\end{equation}

Forgetting is measured as the degradation of old-class performance relative to the initial model:

\begin{equation}
M_f^{(s)}= A_{\mathrm{old}}^{(0)} - 
A_{\mathrm{old}}^{(s)}.
\end{equation}

We report the final forgetting across stages. Lower $M_f$ indicates better retention of previously learned classes \citep{vaze2022generalized,wen2023parametric,wu2023metagcd,park2024online}.

%% file: sections/app_dataset.tex
\section{Datasets and Preprocessing Details}

\label{app:data}

\subsection{Overview}

We evaluate \projectname{} on six public datasets spanning motion capture, wearable inertial sensing, multimodal body sensing, and video-derived human pose. Importantly, these datasets also cover substantially different behavioral domains, including controlled human motion (MoCap), everyday physical and daily-living activities (PAMAP2 and ActRecTut), in-vehicle activities (Drive\&Act), sensor-rich daily routines (OPPORTUNITY), and industrial workplace activities (LARa). Together, they provide substantial variation in sensing modality, sampling rate, dimensionality, sequence length, participant population, and temporal organization, allowing us to evaluate whether recurring temporal structure can be discovered consistently across heterogeneous domains rather than within a single class setting.

Each recording is treated as an independent continuous temporal stream, and discovery is performed without using class annotations or the number of ground-truth classes during training. Ground-truth labels are used only for evaluation after discovery. Unless otherwise stated, segmentation metrics are computed independently within each recording and then aggregated across recordings, avoiding artificial transitions across recording boundaries. As summarized in Table~\ref{tab:datasets}, the benchmark contains 267 recordings and approximately 7.45 million frames in total.

\begin{table}[h]
\centering
\caption{Summary of the six evaluation datasets. \emph{Rec.} denotes the number of independent recordings over which temporal segmentation is evaluated. \emph{Raw ch.} denotes the dimensionality of the original input signals before method-specific preprocessing; the actual input dimensionality used by each method is reported in Table~\ref{tab:preproc}.}
\label{tab:datasets}
\small
\setlength{\tabcolsep}{4pt}
\begin{tabular}{llrrrrrr}
\toprule
Dataset & Source & Subj. & Rec. & Frames & Rate (Hz) & Raw ch. & Classes \\
\midrule
MoCap        & CMU MoCap, subject 86    & 1  & 9   & 65,611    & ---  & 4   & 9  \\
PAMAP2       & PAMAP2 Protocol          & 8  & 8   & 2,864,056 & 100  & 40  & 13 \\
ActRecTut    & ActRecTut                & 2  & 4   & 200,693   & 42   & 24  & 17 \\
Drive\&Act   & Drive\&Act, OpenPose 3D  & 15 & 29  & 1,222,442 & 30   & 104 & 13 \\
OPPORTUNITY  & OPPORTUNITY UCI          & 4  & 24  & 869,387   & 30.3 & 242 & 5  \\
LARa         & LARa v3, MbientLab IMU   & 8  & 193 & 2,224,452 & 100  & 30  & 8  \\
\midrule
\textbf{Total} &                        &    & \textbf{267} & \textbf{7,446,641} & & & \\
\bottomrule
\end{tabular}
\end{table}

\subsection{Preprocessing and input dimensionality}
\label{app:dims}
We convert all datasets into a common representation consisting of continuous
multivariate recordings, annotation track, and recording boundaries. This provides every method with the same temporal
organization and evaluation interface despite differences in the original data
formats.
All analysis windows are restricted to individual recordings. This prevents artificial transitions
from being introduced by concatenating temporally unrelated sequences. More important, ground-truth annotations are retained strictly as evaluation metadata. All
discovery methods operate without access to class labels, class identities,
or the number of ground-truth classes; annotations are exposed only to the
evaluation procedure after discovery.

The datasets vary substantially in input dimensionality, ranging from 4 to 242
raw channels. Because several classical discovery methods become difficult to
apply directly in high-dimensional spaces, we use dimensionality reduction
where necessary while preserving each method's standard representation
pipeline. Time2State and E2USD include learned representation encoders and
therefore receive the native input channels whenever computationally tractable.
For TICC, AutoPlait, and CLaP, higher-dimensional datasets are projected with
PCA using a transformation fitted once per dataset and shared across all
recordings.

We additionally verify that these dimensionality choices do not drive the
reported differences. For Time2State and E2USD, we repeat the experiments using
the corresponding PCA-reduced inputs and obtain results comparable to those
with native-dimensional inputs; we therefore report the native-input results.
For methods using PCA, we evaluate multiple projection dimensionalities and
verify that increasing the retained dimensionality beyond the selected value
does not yield meaningful improvements. Thus, the selected projections retain
sufficient information for each method rather than imposing an artificially
low-dimensional bottleneck.

OPPORTUNITY contains 242 input channels and is high-dimensional even for the
encoder-based methods; we therefore use the same 54-dimensional PCA projection
for all methods, corresponding to the minimum number of components retaining
90\% of the variance. MoCap is already represented by four channels and
requires no reduction. For PAMAP2, we follow the commonly used subset of nine
$\pm16,g$ accelerometer channels from the hand, chest, and ankle sensors.
Table~\ref{tab:preproc} summarizes the resulting preprocessing settings.

\begin{table}[h]
\centering
\caption{Preprocessing settings for each dataset. \emph{Raw} is the original
channel count; the next three columns give the input channels each method
receives (native, the published channel selection, or the number of PCA
components). The window/step applies to the sliding-window methods.}
\label{tab:preproc}
\small
\setlength{\tabcolsep}{4pt}
\begin{tabular}{lrrrrl}
\toprule
Dataset & Raw & Time2State, E2USD & TICC, AutoPlait & CLaP, CompILE & Window / step \\
\midrule
MoCap       & 4   & 4 (native)    & 4 (native)    & 4 (native)    & 256 / 50 \\
PAMAP2      & 40  & 9 (selection) & 9 (selection) & 9 (selection) & 200 / 100 \\
ActRecTut   & 24  & 24 (native)   & 10 (PCA)      & 8 (PCA)       & 90 / 30 \\
Drive\&Act  & 104 & 104 (native)  & 16 (PCA)      & 16 (PCA)      & 90 / 30 \\
OPPORTUNITY & 242 & 54 (PCA)      & 54 (PCA)      & 54 (PCA)      & 30 / 15 \\
LARa        & 30  & 30 (native)   & 16 (PCA)      & 16 (PCA)      & 200 / 50 \\
\bottomrule
\end{tabular}
\end{table}

%% file: sections/app_method.tex
\section{Implementation Details}
\label{app:method}
We provide additional implementation details for the \projectname{}
pipeline used throughout the paper. Unless otherwise noted, all hyperparameters
are fixed across datasets. Class annotations are never used to optimize any
component of the model and are introduced only after motif discovery for
post-hoc semantic mapping and evaluation.

\subsection{Temporal Representation Encoders}
\label{app:encoders}
We used three temporal representation encoders in evaluation: TS2Vec, Time-MAE, and
MantisV2. Each encoder converts a sliding window into a fixed-dimensional
representation that is subsequently processed by UnitAlign. The encoder
parameters are not updated during subsequent static discovery or continual
vocabulary adaptation.Unless stated otherwise, we follow their respective official
implementations and default architectures.

\paragraph{TS2Vec.}
We train a separate TS2Vec model from scratch for each recording. We set the
output representation dimension to 128, use a batch size of 64, and train for
100 optimization iterations. The resulting 128-dimensional embedding is used
as the input to UnitAlign.

\paragraph{Time-MAE.}
We train a separate Time-MAE model from scratch for each recording. Each input
channel is standardized using statistics computed over all windows and time
steps within that recording. We use a patch size of 2 samples, a four-layer
Transformer encoder, a two-layer regressor, dropout 0.1, and batch size 64.
Models are trained for 10 epochs. At inference, the encoder tokens are
mean-pooled to obtain one 64-dimensional representation per window.

\paragraph{MantisV2.}
We use the publicly released pretrained MantisV2 checkpoint without
target-domain fine-tuning or external input normalization. Each window is
linearly resized to length 512 before encoding. For feature extraction, we use
the third Transformer layer (\texttt{return\_transf\_layer=2}) with the
combined CLS and mean-token representation, yielding a 512-dimensional
embedding for each univariate input. For multivariate data, channels are
encoded independently and their embeddings are averaged to obtain one
512-dimensional representation per window. 

Although the encoders produce representations of different dimensionalities, UnitAlign maps all encoder outputs into the same latent space before subsequent discovery. Therefore, the downstream model capacity is held fixed across encoder variants.

\subsection{UnitAlign Implementation}
\label{app:unitalign}
UnitAlign adapts the representations produced by the temporal encoder to the
recurring local structure of the target stream. Let
$y\in\mathbb{R}^{D}$ denote an encoder representation. We use a residual
bottleneck adaptor
\begin{equation}
h =
\mathrm{LayerNorm}
\left(
y +
\alpha
W_2\left(
\mathrm{Dropout}
\left(
\mathrm{GELU}(W_1y)
\right)
\right)
\right),
\end{equation}
followed by
\begin{equation}
z = \mathrm{normalize}(W_hh).
\end{equation}
$W_1$ projects the encoder representation into a bottleneck
and $W_2$ maps it back to the encoder dimension. We initialize $W_2$ to zero
so that the residual adaptor starts close to an identity transformation.
A final projection $W_h$ maps the adapted features to a shared
$\ell_2$-normalized embedding space.

\paragraph{Geometry-preserving initialization.}
Before introducing discrete unit assignments, we first train the adaptor to
preserve the geometry of the encoder representation while avoiding
representation collapse. The objective is
\begin{equation}
\mathcal{L}_{\mathrm{init}}
=
\lambda_{\mathrm{geom}}\,\mathcal{L}_{\mathrm{geom}}
+
\lambda_{\mathrm{VIC}}\,\mathcal{L}_{\mathrm{VIC}},
\end{equation}
where $\mathcal{L}_{\mathrm{geom}}$ preserves the pairwise geometry of the
original representation space:
\begin{equation}
\mathcal{L}_{\mathrm{geom}}
=
\frac{1}{B^2}
\left\|
S(Y)-S(Z)
\right\|_F^2,
\end{equation}
with $S(\cdot)$ denoting the pairwise cosine-similarity matrix within a
minibatch. $\mathcal{L}_{\mathrm{VIC}}$ combines a variance hinge that
discourages dimension-wise collapse with an off-diagonal covariance penalty
that reduces feature redundancy, both applied to the pre-projection adapted
features.

\paragraph{Iterative unit refinement.}
After initialization, UnitAlign alternates between unsupervised clustering and
representation refinement. At iteration $k$, DeepDPM is fitted to the current
adapted representations $z_i$ to obtain pseudo unit assignments
$\hat u_i$. The adaptor is then refined using
\begin{equation}
\mathcal{L}_{\mathrm{refine}}
=
\lambda_{\mathrm{sup}}\mathcal{L}_{\mathrm{SupCon}}
+
\lambda_{\mathrm{geom}}\mathcal{L}_{\mathrm{geom}}
+
\lambda_{\mathrm{VIC}}\,\mathcal{L}_{\mathrm{VIC}}.
\end{equation}

The supervised contrastive term treats windows assigned to the same discovered
component as positives, using only DeepDPM assignments as pseudo-labels and no
semantic annotations. Samples identified as noise by the clustering model are
excluded. The variance and covariance regularizers are applied to the
pre-projection adapted features, while the geometry term continues to preserve
the structure of the original encoder space.

Clustering and adaptor refinement are alternated until the discovered unit
assignments stabilize, as measured by the ARI between
consecutive iterations, or until the maximum number of refinement rounds is
reached. 
After refinement, DeepDPM is fitted once more to the final adapted
representations to obtain the unit vocabulary. Each retained component defines
a local unit, represented by the mean adapted representation of its members.
The resulting unit assignments are temporally smoothed before each window is
replaced by its corresponding unit centroid, forming the centroid timeline used
for motif discovery.

\paragraph{Temporal smoothing.}
The unit assignments produced by DeepDPM are further refined with Viterbi
decoding to discourage implausibly frequent transitions between neighboring
windows. The emission cost is computed in the DeepDPM latent space from the
distance between each window representation and the corresponding unit center,
while a transition penalty discourages changes in unit identity across adjacent
windows. The transition scale is determined from the typical embedding change
between consecutive windows. The resulting smoothed unit sequence is used to
construct the centroid timeline for subsequent motif discovery. 

\subsection{MotifFormer Implementation}
\label{app:motifformer}

MotifFormer operates on the centroid timeline produced by UnitAlign and learns
representations of recurring compositions of local units across multiple
temporal scales. For a subsequence, each unit
centroid is first projected into the Transformer feature space. We prepend a
learnable \texttt{[CLS]} token and add both positional and subsequence-length
embeddings. The resulting sequence is processed by a Transformer encoder, and
the output of the \texttt{[CLS]} token is used as the descriptor of the complete subsequence.

\paragraph{Multi-view motif representation learning.}
To encourage descriptors to capture the underlying recurring composition
rather than individual units, we construct two corrupted views of
each subsequence by masking and dropping unit tokens. MotifFormer is trained
with the objective
\begin{equation}
\mathcal{L}_{\mathrm{motif}}
=
\lambda_{\mathrm{rec}}\mathcal{L}_{\mathrm{rec}}
+
\lambda_{\mathrm{view}}\mathcal{L}_{\mathrm{view}}
+
\lambda_{\mathrm{cross}}\mathcal{L}_{\mathrm{cross}}
+
\lambda_{\mathrm{pseudo}}\mathcal{L}_{\mathrm{pseudo}}
+
\lambda_{\mathrm{var}}\mathcal{L}_{\mathrm{var}}.
\end{equation}

$\mathcal{L}_{\mathrm{rec}}$ reconstructs the original unit-centroid
representations at corrupted positions from the contextualized Transformer
features. $\mathcal{L}_{\mathrm{view}}$ is a contrastive objective that aligns
the \texttt{[CLS]} descriptors of two corrupted views of the same subsequence.
$\mathcal{L}_{\mathrm{cross}}$ further aligns subsequences that begin at the
same temporal position but span different lengths, encouraging MotifFormer to
capture structure that is consistent across temporal scales.

To introduce coarse structural supervision without semantic labels,
$\mathcal{L}_{\mathrm{pseudo}}$ predicts periodically refreshed clusters of
subsequence descriptors. These pseudo-labels are obtained entirely from the
current representation space; only high-confidence assignments are retained
for this term. Finally, $\mathcal{L}_{\mathrm{var}}$ prevents collapse by
encouraging sufficient feature variation across a minibatch. 

\paragraph{Soft coverage across temporal scales.}
Motif discovery proceeds from longer to shorter subsequence lengths. After
motifs at a given length have been identified, their accepted intervals are
recorded on the timeline and used to reweight candidates at shorter scales.
For a candidate subsequence $q$, we compute its coverage ratio
\begin{equation}
\rho(q)
=
\frac{
\left|\mathcal{I}(q)\cap\mathcal{C}\right|
}{
\left|\mathcal{I}(q)\right|
},
\end{equation}
where $\mathcal{I}(q)$ denotes the temporal interval spanned by $q$ and
$\mathcal{C}$ is the union of intervals already claimed by longer motifs.
Its sampling weight is then
\begin{equation}
w(q)=1-\rho(q).
\end{equation}
Thus, a completely uncovered candidate retains full weight, whereas candidates
that largely overlap previously discovered structure contribute less to
clustering without being removed entirely.

\paragraph{Motif matching and multi-length fusion.}
At inference time, the smoothed unit sequence is converted into the same
centroid timeline used during motif discovery. MotifFormer then encodes all
candidate subsequences over the supported temporal lengths. For each candidate,
we evaluate its compatibility with every motif discovered at the corresponding
length.

A candidate is accepted only when it satisfies both a distance and a posterior
criterion in the DeepDPM latent space. Specifically, its distance to the motif
center must fall within the motif-specific acceptance radius, and its posterior
probability under the corresponding mixture component must exceed the stored
confidence threshold. Both statistics are estimated from the motif members
observed during discovery.

For an accepted match between candidate $q$ and motif $m$, we define a
normalized distance
\begin{equation}
r(q,m)
=
\frac{d(q,\mu_m)}{R_m},
\end{equation}
where $\mu_m$ and $R_m$ denote the latent center and acceptance radius of motif
$m$, respectively. Smaller values therefore indicate stronger matches. Within
each subsequence length, if multiple candidates associate the same window with
the same motif, we retain only the match with the smallest normalized distance.

Predictions from different temporal lengths are then fused at the motif level.
Let $\mathcal{L}_{t,m}$ denote the set of lengths that support motif $m$ at
window $t$, and let $r_{t,m}^{(L)}$ be the best normalized distance contributed
by length $L$. We compute
\begin{equation}
s_{t,m}
=
\left(
\frac{1}{|\mathcal{L}_{t,m}|}
\sum_{L\in\mathcal{L}_{t,m}}
\exp\left[
-\min\left(r_{t,m}^{(L)},\tau_r\right)
\right]
\right)
\cdot
b\!\left(|\mathcal{L}_{t,m}|\right),
\end{equation}
where $\tau_r$ limits the influence of weak matches and
$b(\cdot)$ provides a bounded bonus when the same motif is supported by
multiple temporal lengths.
The motif with the highest fused score is selected for each window. Windows
that are not accepted by any motif at any temporal length are assigned to
\textsc{Other}. 

\subsection{Continual Update Implementation}
\label{app:continual}

During continual discovery, each incoming window is first encoded using the
existing temporal encoder and UnitAlign adaptor. 
At the unit level, the incoming window is matched against the existing unit
vocabulary using the same distance-radius and posterior-confidence criteria
described above. If the window is accepted by an existing unit, its
representation is used to incrementally update the corresponding unit
prototype and distribution statistics.

The updated unit timeline is then used to construct subsequences ending around
the newly arrived window. These subsequences are encoded by MotifFormer and
matched against the existing motif vocabulary using the motif-specific
distance and posterior criteria. Accepted
subsequences update the corresponding motif prototype and its matching
statistics.

Data that cannot be confidently explained by the current vocabulary
are not immediately assigned to a new entry. Instead, rejected windows at the
unit level and rejected subsequences at the motif level are placed into
separate novelty buffers. This separation allows the model to accumulate
evidence for persistent new structure before modifying either vocabulary.

\paragraph{Micro-cluster evolution.}
Each novelty buffer is organized as a set of online micro-clusters. A rejected point is either absorbed by a nearby
micro-cluster or used to initialize a new one. Each micro-cluster maintains a
prototype and an effective support that is updated as new points arrive
and decays over time, allowing transient or isolated novelty to disappear
without entering the vocabulary.

A micro-cluster becomes eligible for promotion only after accumulating
sufficient support and exhibiting persistent temporal recurrence. For unit
novelty, candidate points are further consolidated in the existing
DeepDPM latent space so that promotion is based on coherent structure rather
than individual rejected windows. We additionally require the candidate to
span a sufficient temporal extent and to recur across multiple temporal
regions, preventing a short local burst from being interpreted as a new unit.

Once a candidate micro-cluster satisfies the promotion criteria, we compare its
prototype with the existing vocabulary. If it remains compatible with a nearby
existing entry, the accumulated data are merged into that entry and its
prototype and distribution statistics are refined. Otherwise, the candidate is
promoted to a new vocabulary entry and appended to the unit vocabulary.

The motif-level novelty buffers follow the same principle independently at
each temporal length. Rejected subsequences accumulate into length-specific
micro-clusters; persistent clusters are compared with existing motifs at the
same scale and are either merged with a compatible motif or promoted as a new
motif. Consequently, both vocabularies can be continuously \emph{refined} when
new points represent variations of known structure and \emph{expanded}
when persistent previously unseen units or motifs emerge.

\subsection{Implementation Hyperparameters}
\label{app:hyperparameters}

We summarize the key hyperparameters used in our experiments in
Table~\ref{tab:rondo_hparams}. 
\begin{table*}[t]
\centering
\caption{Key hyperparameters used in \projectname{}. Unless otherwise stated,
all settings are fixed across datasets.}
\label{tab:rondo_hparams}
\small
\begin{tabular}{lll}
\toprule
Component & Setting & Value \\
\midrule

\textbf{UnitAlign}
& bottleneck / output dimension & 256 / 128 \\
& initialization epochs & 20 \\
& $\lambda_{\mathrm{geom}},\lambda_{\mathrm{VIC}}$
& 0.2 / 0.05 \\
& refinement epochs / max rounds & 3 / 20 \\
& $\lambda_{\mathrm{sup}},\lambda_{\mathrm{var}},
\lambda_{\mathrm{cov}},\lambda_{\mathrm{geom}}$
& 1.0 / 1.0 / 0.04 / 0.25 \\
& SupCon temperature & 0.1 \\
& convergence & ARI $\geq 0.995$, patience 2 \\
& Viterbi transition scale & 2.0 \\
\midrule

\textbf{DeepDPM}
& latent dimension & 32 \\
& AE training epochs & 10 \\
& DP concentration & 1.0 \\
& unit / motif component cap & 30 / 36 \\
& minimum component weight & 0.01 \\
\midrule

\textbf{MotifFormer}
& lengths & 2--10 \\
& Transformer & 1 layer, $d=128$, 4 heads \\
& training epochs & 20 \\
& mask / token-drop probability & 0.15 / 0.10 \\
& $\lambda_{\mathrm{rec}},\lambda_{\mathrm{view}},
\lambda_{\mathrm{cross}},\lambda_{\mathrm{pseudo}},
\lambda_{\mathrm{var}}$
& 0.1 / 1.0 / 0.5 / 0.5 / 0.01 \\
& contrastive temperature & 0.1 \\
\midrule

\textbf{Motif discovery}
& coverage declaration & strongest 10\% \\
& distance / posterior threshold & P95 / P5 \\
\midrule

\textbf{Continual update}
& novel-unit minimum support & $\max(8,Q_{0.25})$ \\
& merge threshold & $1.25\times$ existing radius \\
& micro-cluster radius & bootstrap NN-distance P95 \\

\bottomrule
\end{tabular}
\end{table*}

%% file: sections/app_baselines.tex
\section{Baseline Implementation Details}
\label{app:baselines}
\subsection{Overview}
We compare against six representative methods for unsupervised temporal
segmentation and state discovery. All baselines are run using the authors'
released implementations, preserving their original learning and inference
procedures. We modify only the data-loading and input/output interfaces required
to apply each method to our datasets.

\begin{table}[h]
\centering
\caption{Baseline methods and their main discovery mechanisms.}
\label{tab:baselines}
\small
\begin{tabular}{ll}
\toprule
Method & Discovery mechanism \\
\midrule
AutoPlait  & MDL-based joint segmentation and regime discovery \\
TICC       & Toeplitz inverse-covariance clustering with temporal smoothness \\
Time2State & causal-CNN representation followed by a DPGMM \\
E2USD      & frequency-aware representation followed by a DPGMM \\
CLaP       & self-supervised classification and segment merging \\
CompILE    & variational learning of segment boundaries and latent codes \\
\bottomrule
\end{tabular}
\end{table}

All methods are converted to a common output format consisting of one predicted
state label per timestamp at the original timeline resolution. Metrics are then
computed once using the same evaluation code for every method. Metrics are defined in
Appendix~\ref{app:metrics}. Each metric is computed separately for every
recording and then averaged across recordings, except SegRatio, which is
computed from the total numbers of predicted and ground-truth segments over the
dataset. We report results on the intersection of recordings for which all
methods produced valid outputs.

\subsection{Experimental Adaptation and Compute}
\label{app:changes}

For AutoPlait, TICC, Time2State, E2USD, DenStream, and DBStream, we use the
authors' released implementations without modifying the underlying algorithms,
using the hyperparameters specified in Appendix~\ref{app:hparams}. DenStream
and DBStream are used only in the continual-discovery experiments and follow
their original online update procedures.
For scale-sensitive methods: \textbf{TICC}, \textbf{CLaP}, and \textbf{CompILE}, we standardize each input channel to zero mean and unit variance within each recording. This avoids differences in channel scale disproportionately affecting TICC's covariance-based state modeling~\citep{hallac2017toeplitz}, CLaP's classification of temporal subsequences~\citep{ermshaus2025clap}, and, in our continuous-input adaptation of CompILE, the reconstruction objective used to learn segment representations~\citep{kipf2019compile}. Per-channel normalization is also a standard preprocessing practice in time-series analysis for reducing sensitivity to offsets and amplitude scales.

\paragraph{CLaP.}
The released implementation requires candidate change points as input rather
than detecting them internally. We obtain these candidates using ClaSP, the
change-point detector on which CLaP is based, and otherwise use the released
implementation unchanged.

\paragraph{CompILE.}
The released implementation is designed for discrete symbolic sequences and
therefore requires minor adaptation to our continuous multivariate streams.
We replace its discrete embedding with a linear projection and use squared
error instead of cross-entropy for reconstruction, while leaving its boundary
inference and latent-variable model unchanged. Long recordings are divided
into 2,000-frame sequences because CompILE uses a fixed segment budget per
sequence and processing entire long recordings is impractical. Finally, we
cluster the resulting segment-level latent representations with a single
dataset-wide Dirichlet process mixture so that state identities are consistent
across recordings, following the clustering protocol used by Time2State and
E2USD.

All experiments ran on a single machine with GPUs (NVIDIA RTX A6000 and A40). Time2State, E2USD and CompILE use a GPU;
AutoPlait, TICC and CLaP are CPU-bound.

\subsection{Hyperparameter Selection}
\label{app:hparams}

No hyperparameter is selected using evaluation performance. We follow the
released configurations and dataset-specific settings of each baseline where
available, and otherwise keep settings fixed across datasets. 

\begin{table}[h]
\centering
\caption{Hyperparameters used for the baseline methods. Values follow the
released implementations or their reference experiments unless noted otherwise.}
\label{tab:hparams}
\small
\begin{tabular}{lll}
\toprule
Method & Setting & Value \\
\midrule
AutoPlait  & parameter-free                  & -- \\
\midrule
TICC       & number of states                & 8 \\
           & window size                     & 1 \\
           & switch penalty                  & 600 \\
           & sparsity                        & 0.11 \\
           & max iterations                  & 100 \\
\midrule
Time2State & encoder channels / depth        & 30 / 10 \\
           & reduced size / output channels  & 80 / 4 (9 on PAMAP2) \\
           & kernel size / learning rate     & 3 / $3\times10^{-3}$ \\
           & $M$ / $N$                       & 20 / 4 (10 / 4 on MoCap) \\
           & training steps                  & 20 (40 on PAMAP2) \\
\midrule
E2USD      & encoder depth                   & 1 \\
           & $M$ / $N$                       & 20 / 4 \\
           & output channels                 & 4 (9 on PAMAP2) \\
           & training steps                  & 20 \\
\midrule
CLaP       & window size                     & SuSS \\
           & classifier                      & ROCKET \\
           & merge score                     & classification gain \\
\midrule
CompILE    & segments per sequence           & 8 \\
           & latent / hidden dimension       & 32 / 64 \\
           & learning rate                   & $10^{-3}$ \\
           & training steps                  & 100 \\
           & sequence length                 & 2,000 frames \\
\bottomrule
\end{tabular}
\end{table}

%% file: sections/app_results.tex
\section{Extra results}
\label{app:seeds}

Table~\ref{tab:full} reports mean ± standard deviation over five seeds for Time2State, E2USD, TICC and RONDO. TICC is deterministic for a given input, so its spread comes only from the seed-dependent PCA fit and is zero on MoCap and PAMAP2, which use no PCA. AutoPlait is deterministic and CLaP uses the fixed random state of its released implementation, so both are run once. CompILE is run once, with seed 0, because of its computational cost.

\begin{table}[!ht]
\centering
\caption{Complete results (\%), mean $\pm$ standard deviation over five seeds.
For SegRatio, values closer to 1 are better.}
\label{tab:full}
\resizebox{\linewidth}{!}{
\begin{tabular}{lccccc@{\hskip 12pt}lccccc}
\toprule
Method & MoF & Recall & F1@50 & Edit & SegRatio & Method & MoF & Recall & F1@50 & Edit & SegRatio \\
\midrule
\multicolumn{6}{c}{\textbf{MoCap}} & \multicolumn{6}{c}{\textbf{Drive\&Act}} \\
\midrule
Time2State & 88.68 $\pm$ 2.07 & 91.49 $\pm$ 2.36 & 84.48 $\pm$ 2.29 & 83.80 $\pm$ 1.17 & 1.70 $\pm$ 0.13 & Time2State & 54.23 $\pm$ 0.39 & 29.55 $\pm$ 1.79 & 7.00 $\pm$ 0.39 & 12.97 $\pm$ 0.47 & 11.31 $\pm$ 0.13 \\
E2USD & 84.97 $\pm$ 2.81 & 87.45 $\pm$ 6.11 & 77.95 $\pm$ 5.77 & 76.48 $\pm$ 6.78 & 1.97 $\pm$ 0.28 & E2USD & 47.60 $\pm$ 0.48 & 18.90 $\pm$ 0.81 & 5.66 $\pm$ 0.29 & 11.80 $\pm$ 0.19 & 14.47 $\pm$ 0.42 \\
TICC & 86.61 $\pm$ 0.00 & 84.52 $\pm$ 0.00 & 79.39 $\pm$ 0.00 & 83.18 $\pm$ 0.00 & 2.19 $\pm$ 0.00 & TICC & 49.47 $\pm$ 0.28 & 23.71 $\pm$ 1.39 & 6.61 $\pm$ 0.65 & 13.01 $\pm$ 0.70 & 9.14 $\pm$ 0.09 \\
AutoPlait & 90.27 & 90.28 & 91.18 & 92.05 & 0.97 & AutoPlait & 39.42 & 8.08 & 0.26 & 4.32 & 0.04 \\
CompILE & 66.45 & 61.05 & 52.19 & 63.14 & 1.97 & CompILE & 46.17 & 17.20 & 9.04 & 32.69 & 4.13 \\
CLaP & 74.61 & 65.63 & 60.53 & 62.39 & 0.74 & CLaP & 49.26 & 17.74 & 11.43 & 20.10 & 0.37 \\
\textsc{Rondo} & 88.91 $\pm$ 1.66 & 96.78 $\pm$ 1.91 & 95.06 $\pm$ 2.37 & 96.09 $\pm$ 2.86 & 2.42 $\pm$ 0.28 & \textsc{Rondo} & 68.36 $\pm$ 0.20 & 62.30 $\pm$ 2.42 & 14.29 $\pm$ 0.63 & 22.22 $\pm$ 0.36 & 8.04 $\pm$ 0.13 \\
\midrule
\multicolumn{6}{c}{\textbf{PAMAP2}} & \multicolumn{6}{c}{\textbf{OPPORTUNITY}} \\
\midrule
Time2State & 79.69 $\pm$ 0.40 & 92.13 $\pm$ 1.87 & 6.17 $\pm$ 0.20 & 8.65 $\pm$ 0.14 & 24.72 $\pm$ 0.48 & Time2State & 69.28 $\pm$ 0.65 & 48.94 $\pm$ 1.07 & 20.80 $\pm$ 1.46 & 58.15 $\pm$ 1.08 & 2.08 $\pm$ 0.05 \\
E2USD & 76.58 $\pm$ 1.48 & 90.21 $\pm$ 2.21 & 4.22 $\pm$ 0.49 & 7.25 $\pm$ 0.43 & 26.96 $\pm$ 1.30 & E2USD & 69.02 $\pm$ 2.21 & 49.95 $\pm$ 2.77 & 26.49 $\pm$ 1.32 & 50.32 $\pm$ 2.20 & 2.91 $\pm$ 0.20 \\
TICC & 74.71 $\pm$ 0.00 & 78.69 $\pm$ 0.00 & 15.93 $\pm$ 0.00 & 16.42 $\pm$ 0.00 & 7.27 $\pm$ 0.00 & TICC & 65.91 $\pm$ 0.80 & 42.99 $\pm$ 1.01 & 13.66 $\pm$ 0.59 & 21.28 $\pm$ 1.29 & 0.37 $\pm$ 0.01 \\
AutoPlait & 45.22 & 29.81 & 21.46 & 21.45 & 0.97 & AutoPlait & 37.64 & 20.83 & 0.00 & 0.63 & 0.01 \\
CompILE & 34.11 & 8.28 & 4.65 & 10.41 & 15.92 & CompILE & 47.13 & 25.54 & 3.44 & 25.24 & 0.46 \\
CLaP & 77.40 & 79.70 & 50.00 & 51.30 & 3.03 & CLaP & 58.88 & 29.61 & 2.82 & 6.61 & 0.09 \\
\textsc{Rondo} & 84.12 $\pm$ 0.93 & 94.25 $\pm$ 1.66 & 22.91 $\pm$ 1.95 & 20.82 $\pm$ 0.80 & 15.37 $\pm$ 0.79 & \textsc{Rondo} & 79.96 $\pm$ 0.27 & 59.19 $\pm$ 0.31 & 30.59 $\pm$ 1.10 & 48.06 $\pm$ 1.73 & 1.23 $\pm$ 0.03 \\
\midrule
\multicolumn{6}{c}{\textbf{ActRecTut}} & \multicolumn{6}{c}{\textbf{LARa}} \\
\midrule
Time2State & 73.40 $\pm$ 2.20 & 49.08 $\pm$ 2.11 & 42.20 $\pm$ 0.73 & 49.07 $\pm$ 2.19 & 1.10 $\pm$ 0.07 & Time2State & 69.91 $\pm$ 0.17 & 37.95 $\pm$ 0.75 & 27.09 $\pm$ 0.83 & 45.89 $\pm$ 0.67 & 2.00 $\pm$ 0.01 \\
E2USD & 72.87 $\pm$ 0.75 & 46.83 $\pm$ 2.44 & 40.41 $\pm$ 2.06 & 45.03 $\pm$ 4.89 & 1.01 $\pm$ 0.04 & E2USD & 67.24 $\pm$ 0.83 & 35.62 $\pm$ 2.08 & 21.81 $\pm$ 1.40 & 40.36 $\pm$ 1.39 & 2.13 $\pm$ 0.02 \\
TICC & 69.68 $\pm$ 0.36 & 41.59 $\pm$ 1.66 & 29.33 $\pm$ 1.02 & 32.24 $\pm$ 1.06 & 0.45 $\pm$ 0.03 & TICC & 70.40 $\pm$ 0.15 & 35.57 $\pm$ 0.69 & 28.34 $\pm$ 1.03 & 44.85 $\pm$ 0.95 & 1.92 $\pm$ 0.02 \\
AutoPlait & 44.42 & 14.06 & 2.04 & 2.31 & 0.01 & AutoPlait & 57.46 & 17.22 & 1.36 & 5.06 & 0.04 \\
CompILE & 45.69 & 27.96 & 8.76 & 31.11 & 0.44 & CompILE & 60.03 & 31.30 & 11.20 & 26.13 & 0.57 \\
CLaP & 51.68 & 27.70 & 5.45 & 5.45 & 0.02 & CLaP & 65.28 & 23.93 & 7.47 & 10.76 & 0.16 \\
\textsc{Rondo} & 75.66 $\pm$ 0.66 & 52.85 $\pm$ 3.34 & 36.71 $\pm$ 2.40 & 45.26 $\pm$ 1.15 & 0.77 $\pm$ 0.03 & \textsc{Rondo} & 78.44 $\pm$ 0.18 & 53.01 $\pm$ 0.78 & 38.50 $\pm$ 0.77 & 47.34 $\pm$ 0.51 & 1.31 $\pm$ 0.01 \\
\bottomrule
\end{tabular}
}
\end{table}